\documentclass{wlscirep}

\usepackage[utf8]{inputenc}
\usepackage[T1]{fontenc}
\usepackage{soul, color}
\usepackage{changes}
\usepackage{multirow}
\usepackage[ruled,vlined]{algorithm2e}
\usepackage{colortbl}
\usepackage{tabu}
\usepackage{bm}
\usepackage{array}
\usepackage{booktabs}
\usepackage{comment}
\usepackage{float}
\usepackage{caption}
\usepackage{titletoc}
\usepackage{etoc}
\usepackage{hyperref}
\usepackage{url} 
 \usepackage{ulem} 
\usepackage{lipsum} 
\usepackage{amsmath}
\usepackage{amsthm}
\usepackage{amssymb}
\usepackage[numbers,super,sort&compress]{natbib}
\usepackage{breakcites}
\usepackage{makecell}
\usepackage{listings}

\usepackage{bm} %
\usepackage{lineno}
\usepackage{ragged2e}

\usepackage{zref-xr}
\zxrsetup{toltxlabel}
\zexternaldocument*{supplementary}

\definecolor{customize}{RGB}{127,34,232} 
\definecolor{darkblue}{RGB}{16, 38, 170}

\titlecontents{section}[0em]{\addvspace{2pt}\filright\bfseries}
{\contentspush{\hspace{0.8em}}}
{}{\titlerule*[8pt]{.}\contentspage}

\titlecontents{subsection}[2em]{\addvspace{1pt}\filright}
{\contentspush{\hspace{0.8em}}}
{}{\titlerule*[8pt]{.}\contentspage}
\usepackage{afterpage}
\titlecontents{subsubsection}[4.5em]{\addvspace{0pt}\filright}
{\contentspush{\hspace{0.8em}}}
{}{\titlerule*[8pt]{.}\contentspage}

\usepackage{titlesec}
\titlespacing*{\section}
  {0pt}{1.2em}{0.6em}

\titlespacing*{\subsection}
  {0pt}{1.0em}{0.4em}

\titlespacing*{\subsubsection}
  {0pt}{0.8em}{0.3em}

\theoremstyle{plain}

\title{\LARGE
BrainFuse: a unified infrastructure integrating realistic biological modeling and core AI methodology
}

\author[1, 2, $\dag$]{Baiyu Chen}
\author[3, $\dag$]{Yujie Wu}
\author[1]{Siyuan Xu}
\author[4]{Peng Qu}
\author[4]{Dehua Wu}
\author[4]{Xu Chu}
\author[4]{Haodong Bian}
\author[5]{Shuo Zhang}
\author[1]{Bo Xu}
\author[4, $\ddag$]{Youhui Zhang}
\author[2, $\ddag$]{Zhengyu Ma}
\author[1, 2, 6, 7, $\ddag$]{Guoqi Li}
\affil[1]{Institute of Automation, Chinese Academy of Sciences, Beijing, China}
\affil[2]{Pengcheng Laboratory, Guangdong, China}
\affil[3]{Department of Computing, The Hong Kong Polytechnic University, Hong Kong SAR.}
\affil[4]{
\parbox[t]{\dimexpr\textwidth-0.09em\relax}{
    \hangindent=0.09em\hangafter=1
    Department of Computer Science and Technology, Beijing National Research Center for Information Science and Technology, Tsinghua University, Beijing, China
}}
\affil[5]{Economics and Management School, Wuhan University, Hubei, China}
\affil[6]{Beijing Key Laboratory of General Brain-Inspired Intelligent Large Models, Beining, China}
\affil[7]{Key Laboratory of Brain Cognition and Brain-Inspired Intelligence Technology, Beijing, China}
\affil[$\dag$]{Contribute equally.}
\affil[$\ddag$]{
\parbox[t]{\dimexpr\textwidth-0.09em\relax}{
    \hangindent=0.09em\hangafter=1
    Corresponding author: zyh02@tsinghua.edu.cn (Y.Z.); mzhygood@gmail.com (Z.M.); guoqi.li@ia.ac.cn (G.L.)
}}

\begin{document}
\begin{abstract}
Neuroscience and artificial intelligence represent distinct yet complementary pathways to general intelligence. However, amid the ongoing boom in AI research and applications, the translational synergy between these two fields has grown increasingly elusive--hampered by a widening infrastructural incompatibility: modern AI frameworks lack native support for biophysical realism, while neural simulation tools are poorly suited for gradient-based optimization and neuromorphic hardware deployment. To bridge this gap, we introduce BrainFuse, a unified infrastructure that provides comprehensive support for biophysical neural simulation and gradient-based learning. By addressing algorithmic, computational, and deployment challenges, BrainFuse exhibits three core capabilities: (1) algorithmic integration of detailed neuronal dynamics into a differentiable learning framework; (2) system-level optimization that accelerates customizable ion-channel dynamics by up to \textbf{3,000$\times$} on GPUs; and (3) scalable computation with highly compatible pipelines for neuromorphic hardware deployment. 
We demonstrate this full-stack design through both AI and neuroscience tasks, from foundational neuron simulation and  functional cylinder modeling to real-world deployment and application scenarios. 
For neuroscience, BrainFuse supports multiscale biological modeling, enabling the deployment of approximately \textbf{38,000} Hodgkin–Huxley neurons with \textbf{100 million} synapses on a single neuromorphic chip while consuming as low as \textbf{1.98 W}. For AI, BrainFuse facilitates the synergistic application of realistic biological neuron models, demonstrating enhanced robustness to input noise and improved temporal processing endowed by complex HH dynamics. BrainFuse therefore serves as a foundational engine to facilitate cross-disciplinary research and accelerate the development of next-generation bio-inspired intelligent systems.
\end{abstract}

\flushbottom
\maketitle

\section*{Introduction}

Computational infrastructure serves as the foundational engine for decoding biological intelligence and advancing artificial intelligence (AI)\cite{daviesAdvancingNeuromorphic2021, wangVirtualBrainTwins2025, sevillaComputeTrendsAccross2022, kaplanScalingLawsNeural2020}, propelling both fields toward their shared aspiration of general intelligence\cite{penkeBrainWhiteMatter2012, werbosIntelligenceInTheBrain2009, yamakawaWholeBrainArchitecture2021, summerfieldNaturalGeneralIntelligence2023}. In neuroscience, simulation platforms such as NEURON\cite{hinesNEURONSimulation1997} and NEST\cite{gewaltigNESTNeuralSimulation2007} have provided the necessary tools to capture the intricate  properties of neuronal dynamics grounded in the Hodgkin-Huxley (HH) models\cite{hodgkinQuantitativeDescriptionMembrane1952, hausserHodgkinHuxleyTheory2000, catterallHodgkinHuxleyHeritage2012}. It thereby advances the scientific understanding of the interrelations among neural structure, behavior, and functionality. In parallel, the AI community has established a comprehensive computing infrastructure comprising auto-differentiation engines, gradient-based optimizers, and specific hardware acceleration\cite{martinezSoftwareEngineringForAI2022, silvanoSurveyDeepLearningHardware2025, raneToolsFrameworksMachine2024}.  This integrated ecosystem has catalyzed the modern AI prosperity\cite{krizhevskyImageNetClassification2012, simonyanDeepConvolutionalNetworks2015, vaswaniAttention2017}, supplying the computational power to architect and train massive foundation models, and boosting the capabilities of state-of-the-art Large Language Models (LLMs)\cite{kaplanScalingLawsNeural2020, hoffmannEmpiricalAnalysisCompute2022, chenLandscapeChallenges2024}.

\begin{figure*}[htbp]
    \centering
    \includegraphics[width=0.99 \linewidth]{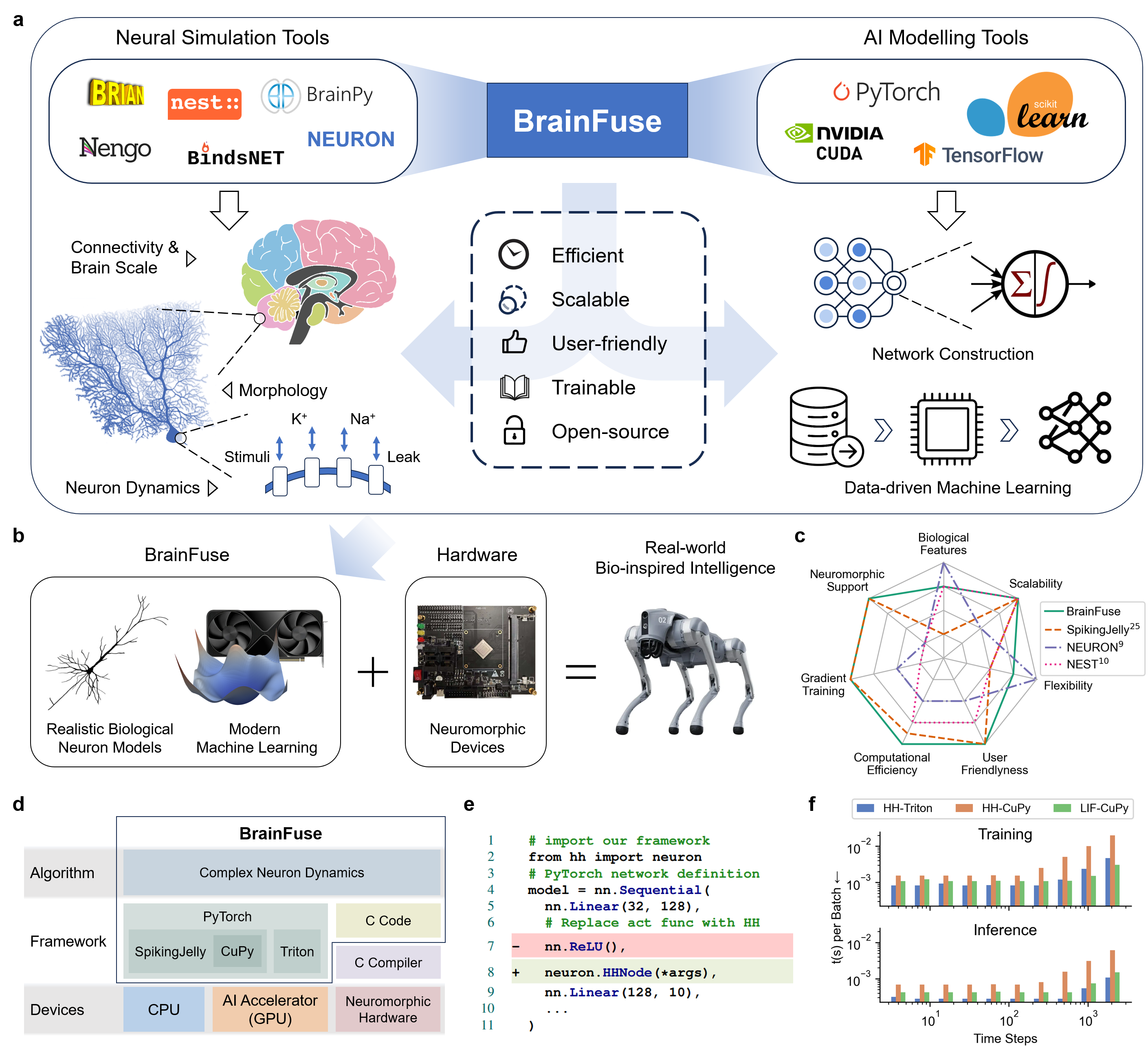}
     \caption{\textbf{Overview of BrainFuse.} 
    (\textbf{a}) Existing neural simulation and AI tooling ecosystems substantially facilitate their respective research domains; however, support for complementary functionalities and their integration remains limited. This gap hinders interdisciplinary research between neuroscience and AI.
    BrainFuse integrates the core functionalities of both sides, providing a user-friendly and efficient interface to model biologically detailed neuron models and train them with standard gradient-based methods. 
    (\textbf{b}) By integrating neuromorphic hardware support, BrainFuse will enable and facilitate the exploration and development of real-world bio-inspired intelligence.
    (\textbf{c}) Comparison of properties between other frameworks for neuron modeling\cite{fangSpikingJellyOpensourceMachine2023, hinesNEURONSimulation1997, gewaltigNESTNeuralSimulation2007} and ours. A detailed comparison is presented in Supplementary Table \ref{table:feat-comp}.
    (\textbf{d}) Architectural design of BrainFuse. The deep integration with modern AI frameworks (PyTorch, Triton) ensures compatibility with a broad range of hardware. A standard C language migration enables BrainFuse to be deployed on customized neuromorphic chips with a C compiler.
    (\textbf{e}) Python code demo to implement a neural network model with HH neurons.
    (\textbf{f}) Computational efficiency comparison between the HH neuron defined in BrainFuse (Triton and CuPy) and the LIF neuron from SpikingJelly\cite{fangSpikingJellyOpensourceMachine2023} (CuPy). We compressed the computational cost of HH to a comparable level with efficient LIF.
}
 \label{fig:summary}
\end{figure*}

Despite these advances, the progress towards bio-inspired intelligence is fundamentally limited by the structural incompatibility between the architectures and computational paradigms underpinning these frameworks (see Fig.~\ref{fig:summary}a). 
Specifically, dominant neuron simulation frameworks have struggled to integrate with modern AI infrastructure\cite{zhangGPUbasedComputationalFramework2023}, particularly GPU-centric computing and gradient-based learning paradigm. 
On the other hand, although the existing AI ecosystem provides a profound basis for gradient-based learning, it lacks native support for the complex neuronal dynamics described by differential equations (DEs)\cite{zhangGPUbasedComputationalFramework2023}, which are foundational to computational neuroscience. 
This mismatch has become a major bottleneck, retarding cross-disciplinary research and the development of advanced intelligence systems\cite{peiArtificialGeneralIntelligence2019, dengEditorialUnderstandingBridging2021}
As a result, much of bio-inspired AI has largely been confined to oversimplified biological models, such as the Leaky-Integrate-and-Fire (LIF) point neuron\cite{fangSpikingJellyOpensourceMachine2023, zengBrainCogSpikingNeural2023, dingNeuromorphicComputingParadigms2025, yaoSpikebasedDynamicComputing2024, choiWaferscaleFabricationMemristive2025}. Meanwhile, contemporary LLMs are constructed upon even more abstract dot-product-based neuron models, missing opportunities to exploit richer, biologically inspired computational primitives.
While several recent studies have begun to explore gradient-based training with complex neuronal dynamics\cite{zhangGPUbasedComputationalFramework2023} or large-scale biological neural network simulations on GPUs\cite{wangBrainScaleEnablingScalable2024, hongSPAIC2024, golosioNestGPURuntimeConstruction2023}, a unified platform that synergizes detailed biophysical simulation, efficient gradient-based learning, and seamless neuromorphic deployment remains lacking.

While highly desirable, establishing such a unified framework is non-trivial, which requires a systematic effort to resolve challenges that span multiple levels.
First, at the algorithm level, enabling gradient-based learning with biologically realistic models requires appropriate discretization strategies for neuronal differential equations and precise gradient derivations for intricate neuronal state updates\cite{fangSpikingJellyOpensourceMachine2023}. Second, at the computing level, efficient training procedures must be implemented on modern AI computing platforms. This requires accordingly refining the given algorithms and performing comprehensive engineering optimization tailored to the specific architecture of modern GPU-like hardware\cite{daoflashattention22023}. Third, at the deployment level, because the architectures of neuromorphic hardware remain divergent and rapidly developing, an adaptive and universal mapping strategy is needed to effectively port these neural algorithms onto these hardware implementations.

To address these challenges, we introduce BrainFuse, \textbf{B}iophysical \textbf{R}esearch and \textbf{AI} \textbf{N}ative \textbf{F}ramework for \textbf{U}nified \textbf{S}imulation and L\textbf{e}arning, a unified computational modeling platform that provides integrated support for fine biological modeling and gradient-based learning. 
Through a holistic architecture co-design spanning algorithm, system, and deployment, BrainFuse enables end-to-end differentiable learning of complex neural dynamics, efficient computation on modern AI infrastructures, and a standard migration procedure onto various neuromorphic hardware platforms. 
To realize these capabilities, we proposed a multi-faceted and integrated development strategy including algorithm refinement, architectural design, and low-level system optimization. First, we developed a refined discretization scheme to reduce the computational cost of biophysical equation solving while preserving realistic behavioral patterns. We further derived exact gradient formulations for HH models and accordingly developed efficient PyTorch-based operators \cite{paszkePyTorch2019} and optimized Triton backends\cite{tilletTriton2019} for fine-grained performance tuning, ensuring seamless integration with the modern AI ecosystem and computational efficiency. Finally, through software and hardware co-design, we migrated the core operators in BrainFuse to C-based implementations to ensure broad compatibility across diverse neuromorphic hardware.

BrainFuse demonstrates three remarkable capabilities through extensive validations on AI and neuroscience tasks. First, it enables multiscale, biologically realistic modeling with up to 3,000$\times$ GPU acceleration compared with native PyTorch implementations while preserving full model expressivity (detailed data illustrated in Supplementary Fig.~\ref{fig:bench_hh_t}). 
Second, it integrates detailed neuronal dynamics into a differentiable learning workflow powered by PyTorch. Through comprehensive sequential learning experiments, we demonstrated the practicality of applying realistic biological neuron models in standard gradient-based learning tasks, elucidating how these intrinsic dynamics inherently augment robustness against input noise.
Third, it supports scalable neuromorphic deployment, enabling cortical-scale networks with approximately 38,000 neurons and 100 million synapses to run on a single neuromorphic chip at power levels as low as 1.98\,W.
 
Together, these results indicate the capacity of BrainFuse to support a complete workflow from neuron simulation, data-driven training, to real-world on-chip deployment, as shown in Fig.~\ref{fig:summary}b. It thereby serves as a powerful tool to bridge the fundamental infrastructure gap between neuroscience and AI, facilitating cross-disciplinary research and accelerating the development of next-generation bio-inspired intelligent systems.

\section*{Results}

\subsection*{BrainFuse: a synergistic platform for neuroscience and AI cross-research}

BrainFuse is designed to integrate realistic biological neuron modeling with differentiable learning support, which is crucial for interdisciplinary research across neuronal simulation and gradient-based learning. Achieving this integration requires breaking the trade-off constraint among biological details, computational efficiency, and operational cost inherent in existing platforms, \cite{fangSpikingJellyOpensourceMachine2023} (see Fig.~\ref{fig:summary}c for the detailed comparison of their core functionalities), which can be traced to three main causes. 
First, detailed neuronal models are governed by a coupled system of differential equations that require complex solving procedures, whereas modern learning frameworks are built around highly abstracted models to simplify numerical optimization. Connecting these two worlds requires careful discretization and stable gradient derivation, which most existing tools do not natively support 
\cite{zhangGPUbasedComputationalFramework2023}. 
Second, most current tools are designed for scientific tasks within domain-specific fields\cite{hinesNEURONSimulation1997, gewaltigNESTNeuralSimulation2007, stimbergBrain2IntuitiveEfficient2019, fangSpikingJellyOpensourceMachine2023}, failing to provide extensible functional support to bridge cross-disciplinary research. Third, most tools lag behind the rapid evolution of modern AI computing infrastructure, preventing them from fully leveraging the computational acceleration provided by GPU clusters to conduct large-scale data mining. 
BrainFuse overcomes this long-standing trade-off between complexity and efficiency through the following three core designs. 

\begin{itemize}

\item \textbf{AI-integrated architecture}.
BrainFuse combines both computationally efficient and user-friendly features in simulation and learning by adopting an AI-integrated architecture. 
To avoid the traditional trade-off between efficiency and engineering complexity, we took a principle of decoupling algorithmic design from low-level technical implementation\cite{ragan-kelleyDecouplingAlgorithms2012}. Recent advances in AI infrastructure have provided a robust foundation for building efficient and user-friendly computing frameworks\cite{paszkePyTorch2019, tilletTriton2019}. Accordingly, BrainFuse is designed to deeply integrate with the modern AI infrastructure (Fig.\ref{fig:summary}d). It enables the convenient construction of AI models that incorporate biological neuron models and customized neuronal dynamics with little loss of efficiency (see Supplementary Information \ref{ssec:code_example}). In addition, it ensures maintainability and extensibility (Code demo in Fig.~\ref{fig:summary}e), which are essential for sustaining an active and productive research community.

\item \textbf{Comprehensive optimization}.
We exploited comprehensive GPU-specific optimization to improve the computational performance of detailed biological neurons, accelerating training and inference by up to 3,000$\times$ over native PyTorch implementations, achieving speeds comparable to simpler LIF models (Fig.\ref{fig:summary}f). Benchmark results are visualized in Supplementary Fig.~\ref{fig:bench_hh_t} and \ref{fig:bench_hh_long}.
This leap in efficiency is realized through the integration of operator fusion, recomputation, and polynomial approximation, allowing for more exhaustive utilization of GPU architecture than previously possible\cite{wangBrainScaleEnablingScalable2024, deistlerJaxleyDifferentiable2025} (see detailed description of these methods in Supplementary Fig.~\ref{fig:op_fusion} and \ref{fig:precision_oscillation}).
Based on the proposed acceleration, BrainFuse significantly lowered the computational barriers once inherent to biophysically realistic models, allowing researchers to simulate and train scaled, realistic neural systems to the level of modern AI workloads, meeting the growing demands of data-driven neural discovery.

\item \textbf{Refined discretization scheme}.
To better suit learning demanding scenarios, BrainFuse carefully rebalances biological fidelity and computational complexity. In numerical integration, accuracy and computational cost are inherently coupled:
smaller step sizes can improve numerical precision and stability, but substantially increase the cost of simulating the same biophysical process, which is undesirable in compute-intensive workloads, including scalable gradient-based learning. We therefore adopted a less precise but more easily conducted discretization with improved consistency and stability in large step-size conditions (see Supplementary Section \ref{ssec:discretization}).
We showed that our discretization scheme can preserve most of the complex neuronal dynamics while compressing the computational cost to an acceptable level, thereby enabling broader research approaches, including gradient-based learning (derivation of gradient computations see Supplementary Fig.~\ref{fig:hh_comp_graph}). 
This is very important to facilitate interdisciplinary explorations between neuroscience and the AI society.

\end{itemize}

These algorithmic refinements reduced the computational complexity while preserving the major behaviors of the biological neurons. Furthermore, they enable keeping a general consistency with contemporary programming languages. A detailed comparison between BrainFuse and previous tools is provided in Supplementary Table~\ref{table:feat-comp} and Section \ref{ssec:neural_enginering_teq}. 

Through the above optimizations, we next demonstrate that BrainFuse enables detailed biological neurons and networks to conveniently migrate onto neuromorphic chips using an intermediate language as the bridge. In addition, the integration with Triton\cite{tilletTriton2019}, a low-level GPU programming framework that focuses on acceleration, provides another flexible approach to deploy BrainFuse based on intermediate representations. We further compiled a C-language version of frequently used BrainFuse components and deployed a function-cylinder-scaled cortical network simulation on a single neuromorphic chip with a power consumption as low as 1.98\,W. The energy efficiency feature facilitates the exploration of biological neural systems in real-world, resource-constrained scenarios. 

Based on BrainFuse, we performed multiple gradient-based learning and biological modeling tasks with HH neuron dynamics\cite{hodgkinQuantitativeDescriptionMembrane1952, pospischilMinimalHodgkinHuxley2008} as a representative example of detailed biological neurons. The results indicate that complex neuron dynamics are applicable in these tasks, showing dynamics-induced robustness to input noises and promising temporal representation abilities.

\begin{figure*}[htbp]
    \centering
    \includegraphics[width=0.95 \linewidth]{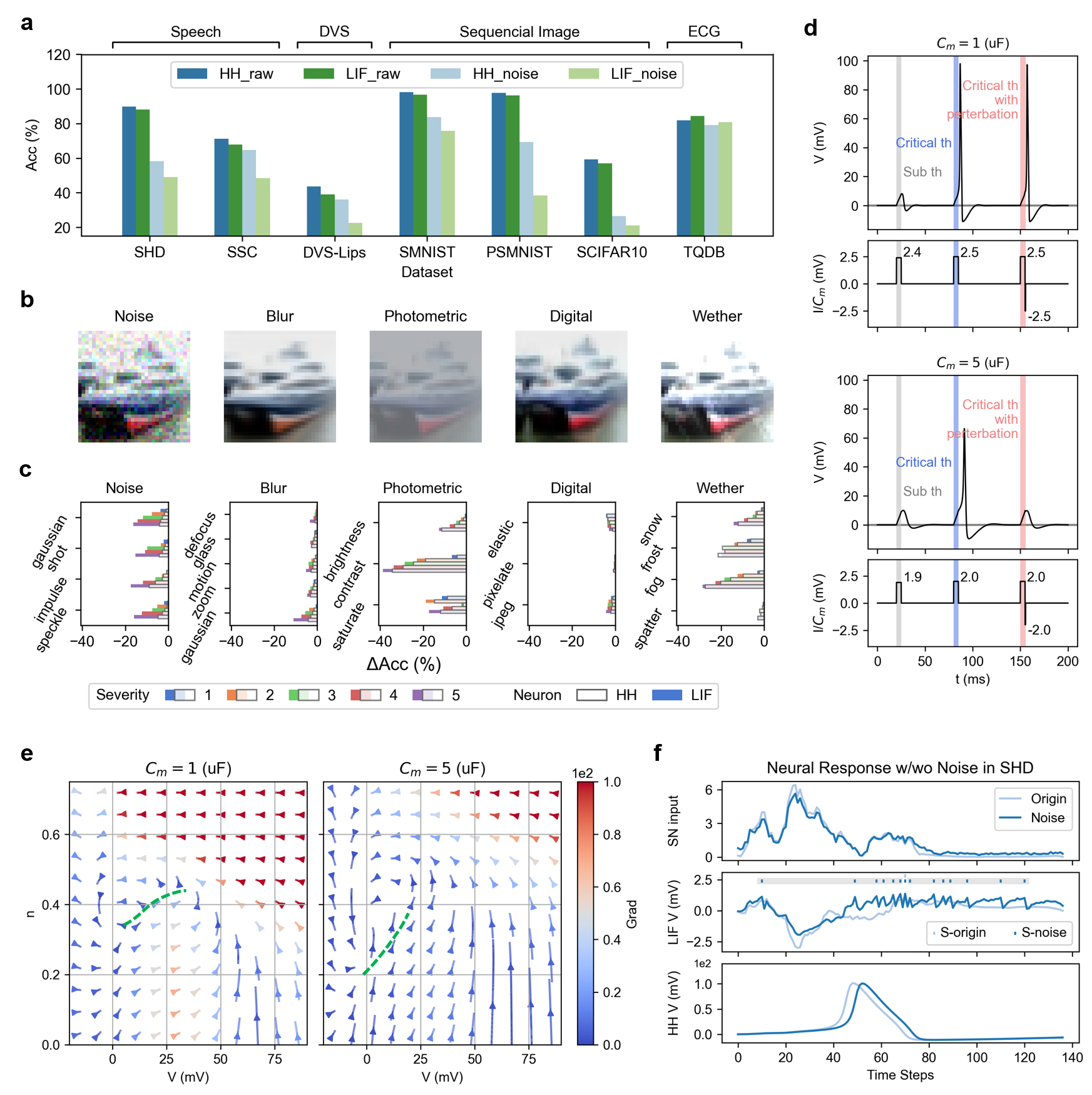}
    \caption{
        \textbf{Robust sequential learning with BrainFuse and analysis of underlying principles.}
        (\textbf{a}) Performance on multiple standard sequential learning tasks of HH models and LIF models, evaluated on clean and noisy test sets. The models are built with BrainFuse and trained on clean training data. The results show that HH neurons generally outperform LIF under both clean inputs and noisy inputs, indicating stronger robustness and representational capacity. A detailed comparison between the refined HH models and advanced Spiking Neural Network (SNN) models is provided in Supplementary Table~\ref{table:ml_results}. The applied noise is task category-specific and described in \textit{Methods}.
        (\textbf{b}) Example images from the CIFAR10-C dataset\cite{hendrycksBenchmarkingNeuralNetwork2019} under severity 4 (amplitude of perturbation, defined in CIFAR10-C dataset).
        (\textbf{c}) Noise robustness comparison on CIFAR10-C. HH exhibits greater robustness than LIF, with lower performance degradation under most corruptions.
        (\textbf{d}) Responses of HH neurons with different membrane capacitances to small perturbations. Neurons with larger membrane capacitance exhibit higher responsiveness. "th" stands for threshold.
        (\textbf{e}) Simplified phase portraits of HH neurons with different membrane capacitances. Neurons with larger membrane capacitance exhibit a steeper separatrix (green dashed lines), making their spiking behavior more susceptible to changes by perturbed input currents.
        (\textbf{f}) Comparison of neuronal behavior in SHD tasks with and without added noise (pepper). The HH neuron’s activity is less affected by input with noise. Spikes of the LIF neuron are indicated by scatter markers (S in the legend), with shaded regions highlighting changed spikes evoked by noise. 
    }
    \label{fig:robust}
\end{figure*}

\subsection*{Efficient gradient-based training for complex dynamics}

The application of realistic biological models in gradient-based learning has remained relatively underexplored, primarily due to the high computational cost of simulating biophysical dynamics and the uncertain advantages offered by such models. With BrainFuse, we (1) significantly compressed the computational cost of conducting gradient-based learning research with HH dynamics and (2) revealed the competence of HH-based models in gradient-based learning tasks, including task performance and robustness to input perturbations. 

We evaluated HH-based models on image classification, speech classification, DVS classification, and time sequence segmentation tasks, then compared their performance with LIF-based models and conventional ANNs under identical training and evaluation setups (see \textit{Methods}). 
Through optimization of both time and memory consumption, the HH-based models with the CuPy backend require only $(1.58 \pm 0.05)\times$ the training time of LIF models, averaged across different hardware environments and learning tasks. The Triton backend further improved the speed by $1.20\times$, which is $1.31\times$ of LIF in estimation (see Supplementary Fig.~\ref{fig:training_time_stat}). HH neurons require only $\sim 1.2\times$ the peak GPU memory usage of a LIF neuron during training, and this difference is expected to decrease when embedded in networks alongside other components. The comparison of memory usage is illustrated in Supplementary Fig.~\ref{fig:mem_comp}.

As shown in Fig.~\ref{fig:robust}a, the HH models achieve comparable or even better than LIF models. This performance advantage is particularly evident in noisy environments.  These results are further corroborated by our comprehensive evaluation on multiple gradient-based learning and neuromorphic benchmarks  (see Supplementary Table \ref{table:ml_results}). 
Figure~\ref{fig:robust}c further shows that the HH models exhibit consistently stronger robustness against diverse input perturbations, including adding noises and image corruptions (visualized in Fig.~\ref{fig:robust}b). 
We attribute this enhanced noise robustness to the intrinsic non-linear dynamics of the HH model. According to the membrane potential evolution Eq.~\ref{eq:hh_dyn_simplified}, in trainable networks, the distribution of input $I$ is adaptive to the firing property of the HH neuron, such that the division by $C_m$ takes a marginal effect on the overall response. In contrast, the non-linear dynamics term $f(V)$ remains unchanged as determined by HH system parameters. Increasing $C_m$ will attenuate the influence of HH intrinsic dynamics while amplifying the influence of external input I, thereby increasing the sensitivity to input perturbations, as shown in Fig.~\ref{fig:robust}d. We further visualize this property from a dynamical system perspective, as shown in Fig.~\ref{fig:robust}e. 
We show in Supplementary Table \ref{table:ml_results},  \ref{table:noise1}, \ref{table:noise2} and  Fig.~\ref{fig:robust_curve} that the above analysis is consistent with the results in learning tasks, where the HH models with larger $C_m$ have higher accuracies with clean input but exhibit more pronounced performance degradation under input perturbations. By contrast, LIF models have no parameter directly corresponding to $C_m$, but the characteristics of their simplified dynamics lead to a similar outcome, that their responses are more severely influenced under input noise (Fig.~\ref{fig:robust}f). 

\begin{figure*}[htbp]
    \centering
    \includegraphics[width=0.99 \linewidth]{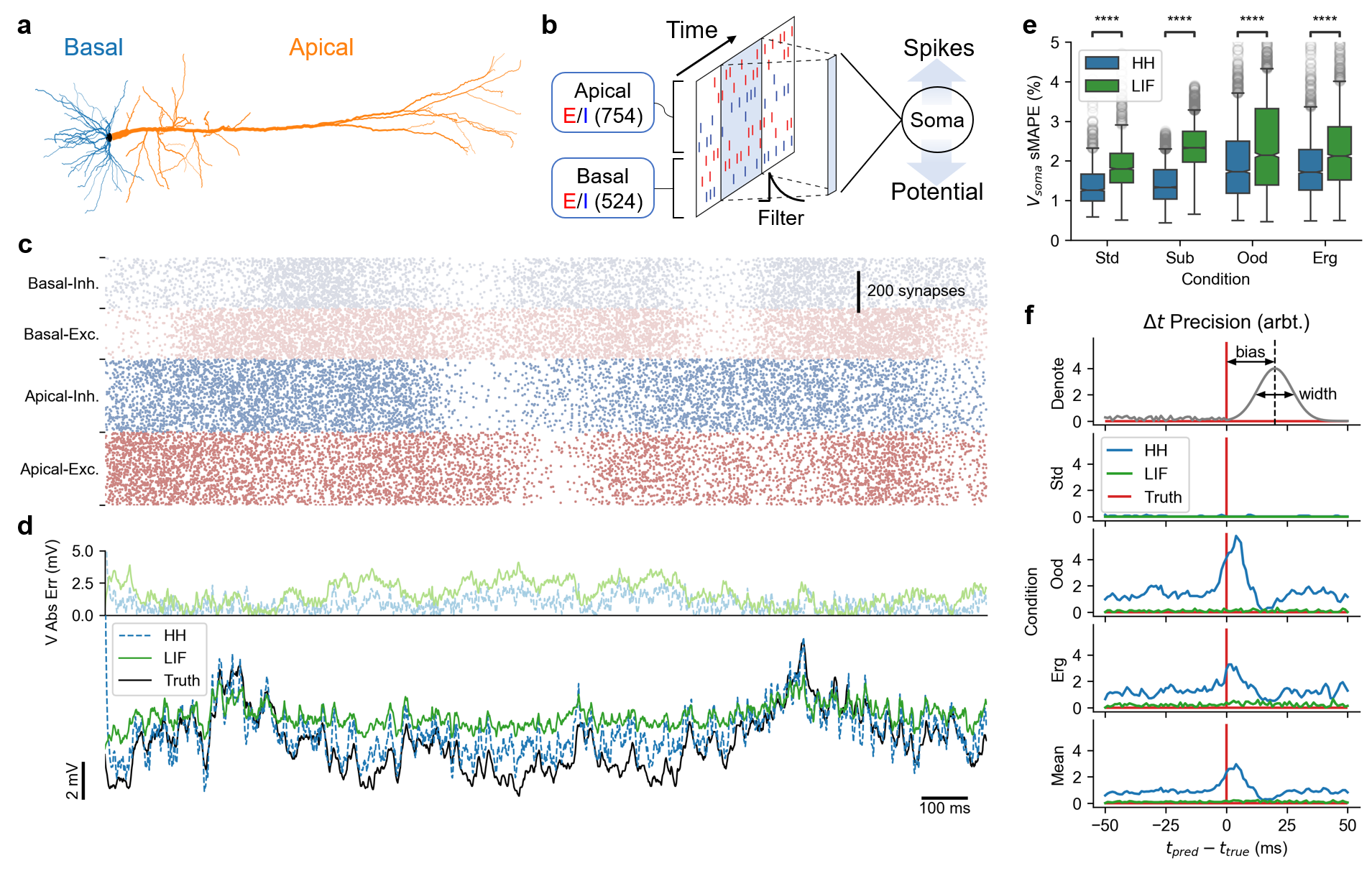}
     \caption{
        \textbf{Neuron scale: L5PC neuron modeling with BrainFuse.} 
        (\textbf{a}) Morphological structure of the L5PC neuron.
        (\textbf{b}) Network architecture designed to model L5PC neuronal activity. Input spikes are convolved with an exponential synaptic kernel and then passed through a fully connected dendritic layer to generate a single-channel somatic input. The point spiking neuron produces both membrane potential and spike outputs (see Methods for detailed setups).
        (\textbf{c}) Visualization of dendritic input spike trains. The y-axis represents different input channels corresponding to distinct synapse types (excitatory/inhibitory).
        (\textbf{d}) Comparison between predicted and simulated membrane potential traces. HH model tracks the ground-truth activities more closely than the LIF model.
        (\textbf{e}) Symmetric Mean Absolute Percentage Error (sMAPE) comparison between HH and LIF models in fitting L5PC membrane potentials under identical architectures and training conditions. In the boxplot, the center line denotes the median, box edges delineate interquartile ranges (IQRs, 0.25-0.75), and whiskers extend from the minimum to the 0.95 maximum values. ****P < 0.0001
        “Condition” refers to the configuration used in NEURON simulations: “Std” for the standard condition, “Sub” for the subthreshold condition, “Ood” for the out-of-distribution condition, and “Erg” for the ergodic condition.
        (\textbf{f}) Distribution of spike-time prediction errors ($t_{\text{pred}} - t_{\text{true}}$) for HH and LIF models, in which y-axis indicates the relative frequency of timing deviations. The “Sub” condition is omitted due to the absence of spikes. The “Denote” panel illustrates the evaluation criteria: a distribution centered near zero (small bias) and sharply peaked (small width) reflects high temporal prediction accuracy and stability. The “Std” condition shows small distribution values due to low spike recall rates: when no spike is predicted within the evaluation window around an actual spike, the case is excluded from spike-time error analysis.
    }
\label{fig:l5pc}
\end{figure*}

\subsection*{Convenient multiscale biological modeling}  

Modeling multiscale biological neural networks, spanning detailed cellular structures to large-scale simulations, is a fundamental but challenging task in biophysical simulation. Early studies typically support only partial modeling capabilities and incur high computational costs.
We addressed this issue in BrainFuse, especially in situations involving large-scale networks and gradient optimization, as described in the earlier sections. Based on our platform, we utilized HH neurons to model realistic biological data at different scales, including (1) single neuron, (2) neural circuit, (3) 1mm\textsuperscript{2} cerebral cortex network, and (4) subcellular morphology.

\paragraph{Neuron scale: L5PC modeling}

For the single neuron scale, we took the Layer 5 Pyramidal Neuron (L5PC)\cite{beniaguevSingleCorticalNeuron2021} as an example, a biologically detailed neuron model with realistic dynamics and morphology, as shown in Fig.~\ref{fig:l5pc}a, simulated with the NEURON simulator\cite{hinesNEURONSimulation1997}. We designed the network structure as Fig.~\ref{fig:l5pc}b, a single linear layer to integrate the input spikes and a scaling layer before the output to align the distribution of membrane potential from different soma models. A E/IPSP decay filter without learnable parameters is applied before the input layer to emulate the dynamics of ion channels. The input is the excitatory and inhibitory spikes from the 639 synapses, as shown in Fig.~\ref{fig:l5pc}c. We trained this model with multiple configurations of supervision, including membrane potential, soma spike, and their linear combination. Detailed configuration is described in \textit{Methods}. 

The goodness of fit can be evaluated by comparing the predicted membrane potential and spike sequences with the ground truths. Figure~\ref{fig:l5pc}d and e indicate that the HH soma unit models the membrane potential with lower error. Statistically, as shown in Fig.~\ref{fig:l5pc}e, HH soma generally has a smaller fit error than LIF in all condition splits. The HH soma has similarly low fit error in Std and Sub split, while the LIF soma has significantly higher loss in Sub than in Std. 
Comparing normal (Std, Sub) and abnormal (Ood, Erg) conditions, Fig.~\ref{fig:l5pc}e shows that both HH and LIF somata have larger IQRs or dispersions in abnormal conditions than in normal, implying the richness of biological neuronal dynamics is more substantially presented under abnormal stimuli. 

Notably, all samples in L5PC's training set are generated under the Ergodic condition (Erg split in test set). In normal conditions, the fit error distribution of LIF largely overlaps with Erg, while HH tends to be lower than Erg. This observation implies that HH aligns its dynamics with the given model, while LIF tends to fit the data statistically.
Fig.~\ref{fig:l5pc}f visualizes the distribution of the spike-time prediction errors, i.e., the difference between the predicted spike time $t_{pred}$ and ground truth $t_{true}$ of each recalled spike. The result shows that HH soma generally recalls more spikes and tends to predict the recalled spikes at the correct timings. A comprehensive analysis of the results (Supplementary Fig.~\ref{fig:firing_patterns} and Fig.~\ref{fig:l5pc_curves}), together with the evaluation methodology for spike prediction, is included in Supplementary Section \ref{ssec:L5PC_setup}. 
These analyses further underscore the necessity of incorporating biophysical mechanisms when modeling the dynamics of biological neurons.

\begin{figure*}[htbp]
    \centering
    \includegraphics[width=0.99 \linewidth]{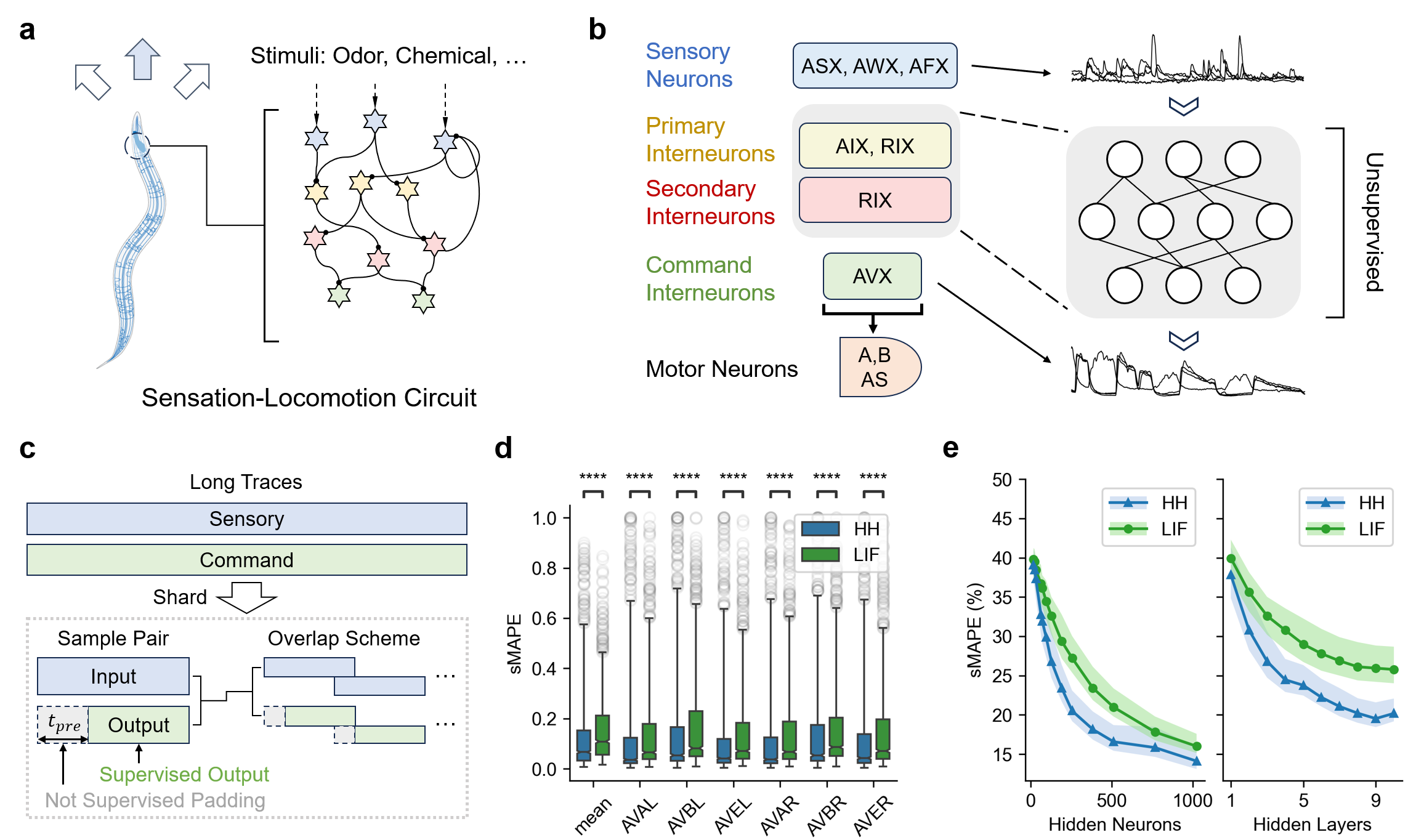}
     \caption{\textbf{Circuit scale: C. elegans sensory-motor circuit modeling with BrainFuse.} 
    (\textbf{a}) Schematic chart of the sensory-locomotion neural circuit.
    (\textbf{b}) The structure of the neural circuit and the modeling scheme. The abbreviations in the boxes represent specific neurons in the C. elegans nervous system, where X represents multiple related neurons (see \textit{Methods} for details). We take calcium imaging traces of the sensory neurons as input, command interneurons as output, and model the functionality of primary and secondary interneurons using an SNN multilayer perceptron (MLP) model.
    (\textbf{c}) Scheme of data preprocessing. We segmented the long traces into shorter shards and preserved an unsupervised padding at the beginning of each sample pair. No supervised output overlaps among samples.
    (\textbf{d}) Comparison of fitting error between HH and LIF, where the x ticks are different output neurons and their average. The HH model consistently tracks the biological recordings more precisely than LIF. In boxplot, the center line is the median, box edges delineate interquartile ranges (IQRs, 0.25-0.75) and whiskers extend from minimum to 0.95 maximum values, ****P < 0.0001.
    (\textbf{e}) Verification of hyperparameter sensitivity. For different model depths and widths, HH achieves lower error than LIF in all configurations. For hidden neuron trials (changing the number of neruons in one hidden layer), the depth is fixed to 3. For hidden layer trials, the width is fixed to 64.
 }
 \label{fig:c_elegans}
\end{figure*}

\paragraph{Circuit scale: C. elegans modeling}
At the neural circuit level, we verified the HH network by modeling the sensory-motor circuit of C. elegans, as shown in Fig.~\ref{fig:c_elegans}a. For this task, we used the data from a collection of C. elegans calcium imaging neural activity datasets\cite{simeonHomogenizedCelegansNeural2025}. A subset of input and output neurons was selected, records with missing traces were removed, and the remaining traces were divided into equal-length segments, which were then randomly assigned to training and testing datasets in a 3:1 ratio. We simply used a spiking neuron multilayer perceptron (MLP) to model the internal dynamics between sensory neurons and command interneurons, as shown in Fig.~\ref{fig:c_elegans}b. To let the spiking neuron prepare the internal state, we included a preparing stage at the beginning of each sample, as illustrated in Fig.~\ref{fig:c_elegans}c. We then trained the networks using LIF and HH neurons and examined the fitting error. 

The result in Fig.~\ref{fig:c_elegans}d shows that HH fits the dynamics of the neural circuit with smaller error. We varied the width and depth of the network to rule out artifacts in hyperparameter selection. As shown in Fig.~\ref{fig:c_elegans}e, HH outperforms LIF in a wide range of model scales. 

Note that in comparison with L5PC fitting (Fig.~\ref{fig:l5pc}e), the sMAPE of C. elegans fitting (Fig.~\ref{fig:c_elegans}d) is significantly larger and heavy-tailed. A possible interpretation is that, in comparison with the adopted HH neuronal dynamics, C. elegans' dynamics is more distinctive than L5PC's. Another explanation is the mismatch of timescales between the behaviors of neuronal activity and Calcium imaging. 
Since this fitting experiment is substantially closer to a black-box fitting than a data-driven modeling, as there is no topological constraint between the actual circuit connections and the fit model, the modeling result is less ideal than L5PC, which has higher structural similarity. This phenomenon encourages stronger structural alignment for better computational modeling.

\begin{figure*}[htbp]
    \centering
    \includegraphics[width=0.92 \linewidth]{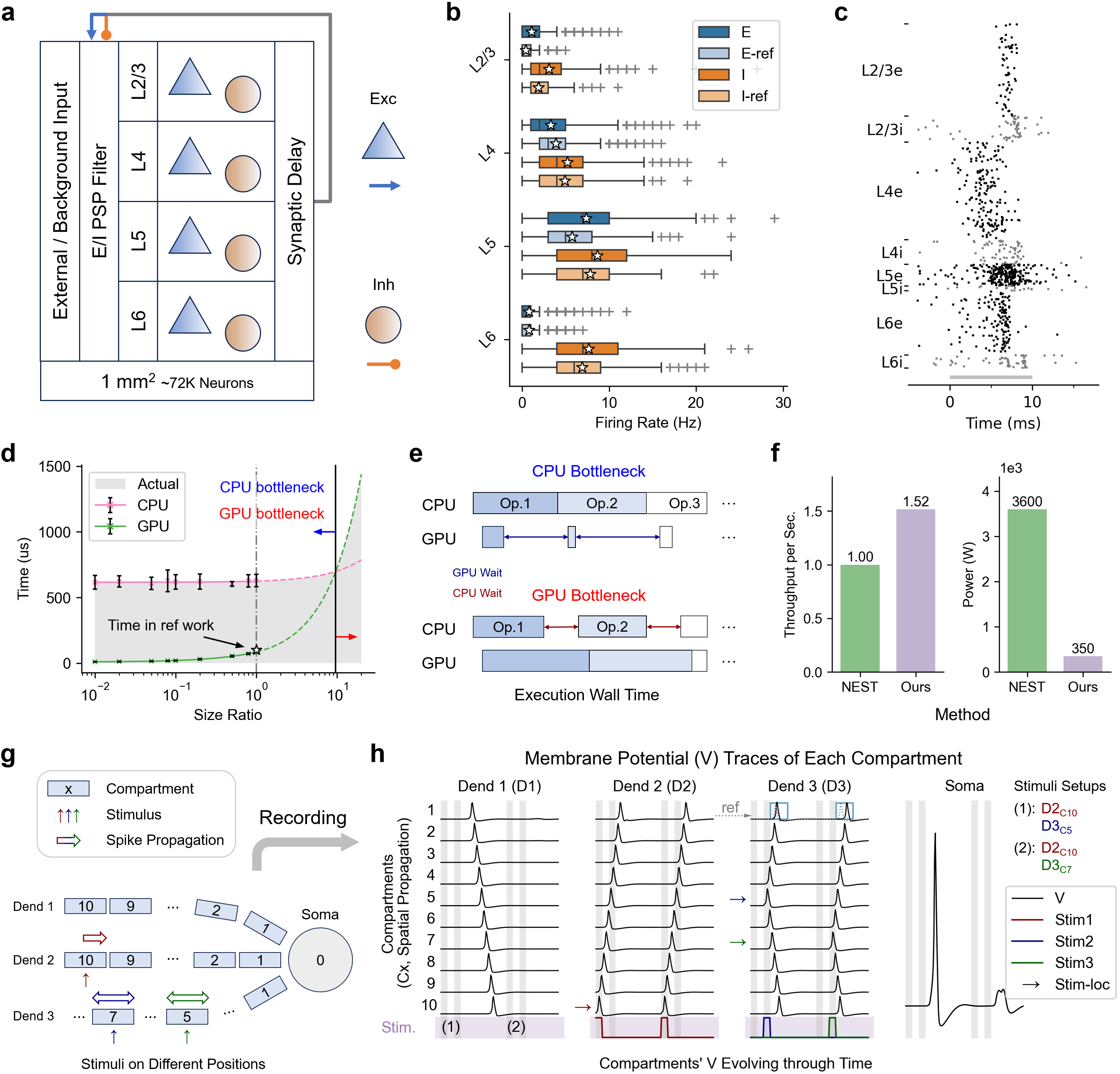}
     \caption{\textbf{Cortical \& subcellular scale: scalability in network connection and neuronal morphology with BrainFuse.} 
    (\textbf{a}) Configuration of the cerebral cortex network simulation. The network is composed of excitatory and inhibitory neurons from 4 cortical layers. These 8 neuron populations are connected using the modeled connectivity probability and latency distributions, described in reference work\cite{potjansCell-type2014}. 
    (\textbf{b}) Comparison of firing rate statistics between the reference work's result and ours, in the resting state simulation. In the boxplot, the center line is the median, star is the average, box edges delineate interquartile ranges (IQRs, 0.25-0.75), and whiskers extend from minimum to 0.95 maximum values.
    (\textbf{c}) Raster plot of model response with transient thalamo-cortical input stimulation, reproduced by our model based on BrainFuse.
    (\textbf{d}) Relation between simulated model size and per-iteration overhead allocation. The actual time cost is marked by the shadow area. GPUs will outperform CPUs at roughly $10\times$ larger model size, indicating promising scalability.
    (\textbf{e}) Diagram of CPU and GPU bottleneck situation. For each executed operation (Op.), the slowest one of the host and devices (GPU) determines the launch time of the next Op. GPU waiting leads to lower utilization and simulation throughput.
    (\textbf{f}) Comparison of power consumption and simulation throughput. The baseline results are estimated from the referenced work, conducting the simulation on a 48-CPU cluster. Our results are based on BrainFuse with a single NVIDIA 3090 GPU. This comparison indicates BrainFuse's advantages of energy-efficiency and batched throughput.  
    (\textbf{g}) Diagram of the spatially discretized neuron morphology and the stimuli scheme.
    (\textbf{h}) Soma responds differently with the same input pattern applied to different positions. The vertical shadows label the simulation timing. Trials (1) and (2) have identical stimulus durations and intervals between the two stimuli. The difference in spike arrival timings at the soma leads to different responses (D2-C1 trace also plotted in D3-C1 panel with gray dotted line to assist timing comparison, denoted by blue boxes). See the animation of this simulation in Supplementary Movie S1.
 }
 \label{fig:brain_sim}
\end{figure*}

\paragraph{Cortical scale: cortex simulation}
To further demonstrate the scalability of BrainFuse, we reproduced the cerebral cortex network simulation \cite{potjansCell-type2014} with our platform. We followed the primary modeling configuration as Potjans' (Fig.~\ref{fig:brain_sim}a), with only the LIF neurons replaced with HH and several parameters modified to adapt to the neuron replacement (see \textit{Methods} and Supplementary Section \ref{ssec:cortex_sim_setup}). The results indicate that we have successfully reproduced the static rest state firing rate distribution and the propagation pattern under transient thalamus stimuli (Fig.~\ref{fig:brain_sim}b and ~\ref{fig:brain_sim}c), as validations of the rationality of our setup.
After correctly reproducing the results, we further applied comprehensive optimizations to exhaust the computing power of modern computing architectures (see Supplementary Section \ref{ssec:cortex_input_simp} and Fig.~\ref{fig:bg_gen_comp}). A properly planned execution process is crucial for an algorithm to run efficiently. To reveal the bottleneck in cortex simulation, we recorded the activities and summarized the active time of CPU and GPU, as shown in Fig.~\ref{fig:brain_sim}d. As the size of the simulated model increases, GPU active time rises significantly, whereas CPU time increases only marginally. The actual execution time depends on the slowest process, predicting that if the model size is further scaled up, the bottleneck would shift from CPU to GPU, as depicted in Fig.~\ref{fig:brain_sim}e. 
To better exploit modern GPU architectures with flexible model size and support gradient-based learning workloads, we introduced the data parallel feature into the cortex simulation model for improving device utilization. 
With an appropriate parallel configuration, we can fully leverage the computational power of GPUs to their maximum.
With the support of our platform, we are able to conduct the simulation on a single NVIDIA RTX3090 24G GPU, achieving a 1.5$\times$ throughput (the product of the number of paralleled simulations and the simulation speed) with substantially  lower energy consumption in comparison with Potjans' work, as shown in Fig.~\ref{fig:brain_sim}f. 

\paragraph{Sub-cellular scale: neuronal morphology modeling}

In addition to the scalability in network connections, BrainFuse also supports scalable neuronal morphology. As an example, we construct a neuron model comprising three dendrites and one soma, as illustrated in Fig.~\ref{fig:brain_sim}g, where the spatial distribution is discretized into compartments. Using this model, we demonstrated the temporal signal integration characteristics of a biologically detailed neuron in Fig.~\ref{fig:brain_sim}h. When two grouped stimuli with the same duration and interval (see (1) and (2) labeled with different color combinations in Fig.~\ref{fig:brain_sim}h) are applied on different compartments of the dendrite, the soma would respond differently according to the arrival time interval between the spikes coming from different dendrites. The spike on Dend1 is evoked by the soma activation in trial (1), without external stimuli.

\begin{figure}[htbp]
    \centering
    \includegraphics[width=1\linewidth]{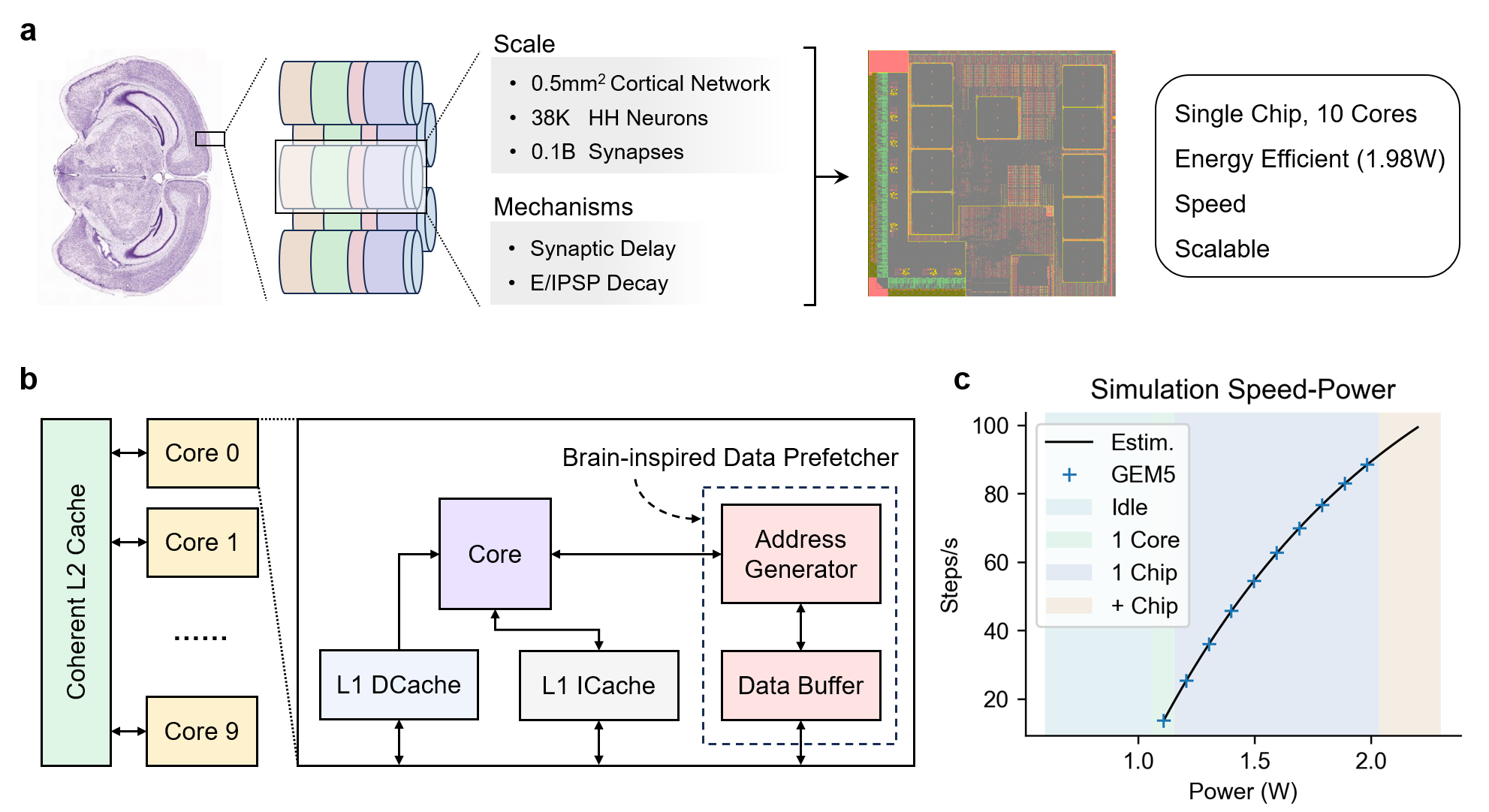}
    \caption{\textbf{BrainFuse deployment on neuromorphic hardware}.
    (\textbf{a}) Deployment of a 0.5mm\textsuperscript{2} cortical network on a single neuromorphic chip. The combination of BrainFuse and the neuromorphic chip can simulate realistic biological neural networks on a single chip at a power consumption as low as 1.98\,W.
    (\textbf{b}) System architecture of the neuromorphic chip. It improves the speed and energy efficiency in indirect addressing memory access scenarios with the specially designed data prefetcher. 
    (\textbf{c}) The relationship between power consumption (activated cores) and simulation speed on the neuromorphic chip. Idle indicates the power of the other compartment except the cores. The data points are generated from GEM5 simulation, an architectural simulator, to validate the functional correctness and performance of chips. We estimated the relationship (solid black line) according to the simulated data.
    }
    \label{fig:thruster}
\end{figure}

\subsection*{Practical deployment on neuromorphic devices}

Achieving energy efficiency is one of the key challenges in developing real-world embodied bio-inspired intelligence\cite{royMachineLearningRoboticsChallenges2021}. Based on the comprehensive optimizations of BrainFuse, various biological neuron, circuit, and network models become deployable on energy-efficient edge devices. We demonstrated the energy-efficient advantage of the system comprising BrainFuse and the neuromorphic hardware with the following on-chip cortical network simulation. 

The biophysical simulation approach provided by BrainFuse is deployable on low-power neuromorphic hardware with the optimized computing schemes. We ported the cortical network simulation model onto a neuromorphic chip as the evaluation of the affinity to neuromorphic hardware (Fig.~\ref{fig:thruster}a). The adopted neuromorphic chip is a new-generation evolution of the GaBAN architecture\cite{chenGaBANGenericFlexibly2022}, as shown in Fig.~\ref{fig:thruster}b. It employs a heterogeneous architecture that fuses bio-inspired computing with general-purpose computing, enabling deep integration of a specialized unit for optimizing sparse memory access (inherent to bio-inspired operations) with the general-purpose chip (see details in Supplementary Table \ref{table:thruster}). 

This architecture demonstrates excellent energy efficiency in indirect memory accessing scenarios, where the cortical network is a typical example with sparse synaptic connections. We successfully simulated a 0.5mm\textsuperscript{2} cortical network with approximately 38,000 HH point neurons and 0.1 billion synapses on a single neuromorphic chip, with an elastic power consumption range between 0.20W and 1.98\,W, as shown in Fig.~\ref{fig:thruster}c. The data of simulation speed and power consumption is generated through GEM5 simulation, a cycle-approximate, event-driven architectural simulator (see details in Supplementary Fig.~\ref{fig:thruster_supp}). Combined with BrainFuse's flexible support for dynamics extension, it is possible to deploy realistic biological neural circuits on embodied agents, connecting neural activities with physical interactions in the environment.

Collectively, these results indicate that BrainFuse, in combination with neuromorphic hardware, constitutes a complete toolchain for exploring biologically detailed neuron models, supporting flexible model development, fast data-driven training, and energy-efficient deployment. BrainFuse is expected to facilitate the progress towards a new stage of bio-inspired intelligence.

\begin{figure*}[htbp]
    \centering
    \includegraphics[width=0.99 \linewidth]{"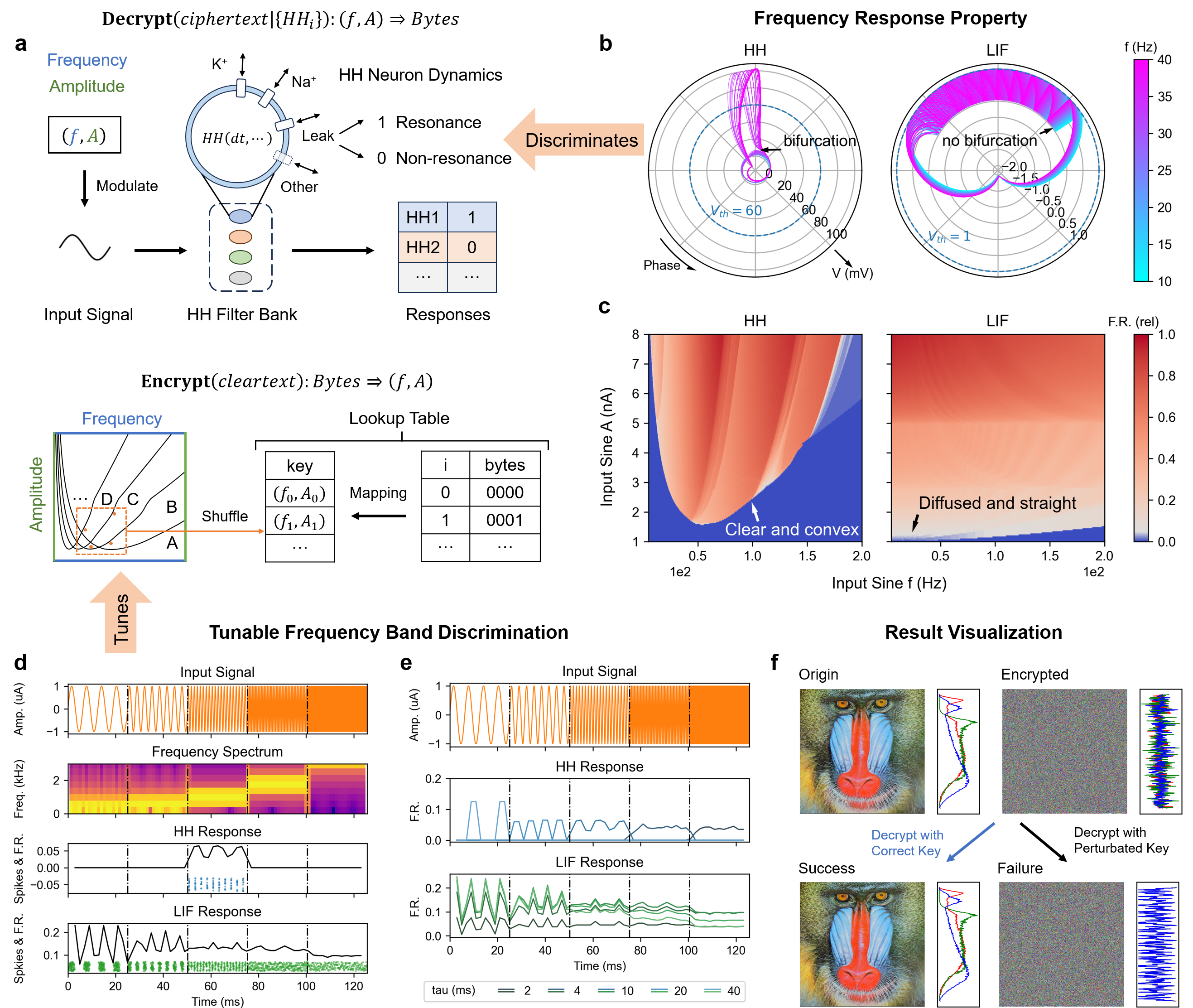"}
     \caption{\textbf{HH-oscillation inspired encryption and decryption with BrainFuse.} 
    (\textbf{a}) Schematics of HH encryption–decryption principle and procedure. 
    Inspired by HH's dynamics visualized in (b) and (c), we designed an HH-based decoding algorithm. A filter bank, composed of multiple HH neurons with different parameters, can generate a multi-bit response according to the frequency of the input sine signal. We then developed a corresponding encoding rule to match the decoding property, and the parameters of the sine waves act as the secret key.
    (\textbf{b}) State trajectories of HH and LIF neurons under sinusoidal stimuli with different frequencies, with the angle as the phase of the sine function in $\pi$ units and the radius as the membrane potential. A frequency-dependent bifurcation is observed in HH, where it transits from fluctuation to periodic spiking. In contrast, LIF exhibits a periodic spiking pattern over a certain threshold, independent of frequency variation.
    (\textbf{c}) Comparison of response characteristics (firing rate) of HH and LIF neurons to sinusoidal signals with varying frequencies and amplitudes. The HH model exhibits a distinct phase transition boundary, whereas the LIF model shows a relatively smooth response. 
    (\textbf{d}) Comparison of frequency selectivity between HH and LIF neurons, demonstrating that HH neurons possess nonlinear frequency selectivity. 
    (\textbf{e}) The frequency selectivity of HH is tunable by modifying the time constant parameter $\Delta t ~ (1/\tau)$ of HH. Conversely, modifying the similar parameter $\tau$ in LIF does not have the same effect.
    (\textbf{f}) Demonstration of image encryption using the principle and procedure described in (a). Origin denotes the original image; Encrypted shows the encrypted image, with close to uniform distribution; Success represents the correctly decrypted image (expected to match the original); and Failure shows the result of decryption using a perturbed public key with added random noise. The corresponding RGB histograms are shown on the right.
 }
 \label{fig:encryption}
\end{figure*}

\subsection*{Accessible exploration of biological neuron models' applications}

BrainFuse eases the extensive exploration of biological neuron models. Biological neuron models, such as HH, exhibit much more complex dynamics than simplified neurons like LIF. However, due to the high barrier to accessing existing tools for neuron dynamics, effectively leveraging these complex dynamics in practical applications remains an open question. Next, we show that BrainFuse provides a user-friendly interface for the community to develop applications based on complex neural dynamics. 

To demonstrate the application potential of complex neurons, we designed an encryption-decryption algorithm using the HH neuron based on the proposed platform, as shown in Fig.~\ref{fig:encryption}a. 
The core principle of this algorithm is that HH dynamics have an inherent oscillatory property\cite{izhikevichDynamicalSystemsNeuroscience2006}. As shown in Fig.~\ref{fig:encryption}b, the HH neuron will shift from silence to a periodically firing state at a certain frequency threshold of the sinusoidal input signal, as the labeled bifurcation point in the left panel, where LIF behaves similarly among all frequencies. 
In Fig.~\ref{fig:encryption}c, we can see the specific response to the input signal as amplitude and frequency vary. HH has a convex and sharp phase-shift border within the frequency-amplitude space (f-A space), where LIF behaves more similarly to a linear threshold. 
The specific response patterns of HH and LIF with different frequencies are further visualized in Fig.~\ref{fig:encryption}d. LIF tends to respond more strongly under low-frequency inputs, where HH respond only when the input frequency falls within a certain range. The responsive range is tunable by varying the time constant $\tau$, one of the parameters of the HH neuron, as shown in Fig.~\ref{fig:encryption}e. 

Following this principle, we designed the encryption algorithm. HH's phase-shift border in f-A space acts as a binary classifier. By varying HH's parameters, we can compose an HH filter bank to achieve combined encoding. As shown in the upper part of Fig.~\ref{fig:encryption}a, the HH filter bank can map an (f,A) pair into a multi-bit binary representation. The response is determined by the parameters of each neuron within the filter bank. Following the HH response pattern, we construct an inverse mapping by selecting specific points in the f-A space and assembling a lookup table that maps byte values to corresponding (f,A) pairs (visualized in Supplementary Fig.~\ref{fig:hh_lif_resp_tau}-\ref{fig:hh_filter_bank_subspaces}). For detailed descriptions of this algorithm, see \textit{Methods} and Supplementary Algorithm \ref{algo:hh-encryption} and \ref{algo:hh-decryption}. 

With the HH neuron provided in BrainFuse, we implemented this algorithm and verified its effectiveness, as shown in \ref{fig:encryption}f. The encrypted image loses the original distribution. Only with exactly correct keys can the image be decrypted.  A tiny perturbation to the keys would lead to decryption failure and completely disrupt the original image distribution, degrading into noise. This experiment demonstrates a convenient exploration of applying biological neuron dynamics in practical scenes. With these accessible components, BrainFuse invites broad participants to exploit the potential applications of biological neuron models.

\section*{Discussion}

We have presented BrainFuse, a platform with high flexibility, user-friendliness, AI affinity, and computational efficiency for realistic biophysical simulation and AI modeling.
A key feature of BrainFuse is its accessible interface, which allows users to customize ion-channel equations and morphologies with nearly minimal programming requirements. Specifically, users need only to define the conductance term, the ion-channel activation and deactivation term, and the corresponding derivatives in a straightforward numpy-like style (detailed code examples provided in Supplementary Code~\ref{code:brainfuse_ion_channel} and \ref{code:brainfuse_morph}). 
Leveraging its auto-differentiation engine and PyTorch-native design, BrainFuse facilitates the seamless incorporation of complex neuronal models into standard deep learning workflows.

Engineering optimization was one of the central focuses in the design of BrainFuse, as it fundamentally dictates the platform's efficiency and real-world applicability.
By integrating comprehensive computational optimization with appropriate algorithmic simplification, BrainFuse adapts seamlessly to modern AI accelerators and neuromorphic hardware.
On GPUs, we demonstrated that BrainFuse reduces the computational cost of simulating complex neural dynamics to a level comparable to the simplified LIF model. For neuromorphic hardware deployment, we demonstrated the feasibility of mapping a biologically detailed neural network onto a single chip with a low power consumption of merely 1.98\,W.

Another prominent hallmark of BrainFuse is its efficient and seamless integration of biologically detailed neuron modeling with gradient-based learning. While seminal platforms have shared ultimate goals, their designs are highly specialized\cite{hinesNEURONSimulation1997, gewaltigNESTNeuralSimulation2007, wangBrainPyFlexibleIntegrative2023, fangSpikingJellyOpensourceMachine2023}.
NEURON\cite{hinesNEURONSimulation1997} focuses on the fidelity to simulate very detailed neuron models, NEST\cite{gewaltigNESTNeuralSimulation2007} can support large-scale network simulations with relatively simple neuron dynamics, and BrainPy\cite{wangBrainPyFlexibleIntegrative2023} provides a friendly user interface for customizing new dynamics. BrainFuse provides scalable capabilities for neuron dynamics and morphology modeling, with additional accessible support for gradient-based learning exploration, which other platforms do not. DeepDendrite\cite{zhangGPUbasedComputationalFramework2023} implemented gradient-based learning with a detailed neuron model, but the learning support is limited to the specific model, making it inconvenient to cooperate with other AI frameworks. 
SpikingJelly\cite{fangSpikingJellyOpensourceMachine2023}, Nengo\cite{bekolayNengoPythonTool2014}, and BrainCog\cite{zengBrainCogSpikingNeural2023} provide efficient spiking neural network construction and convenient training, but they do not support complex neuronal dynamics and detailed morphology modeling. 
While BrainScale \cite{wangBrainScaleEnablingScalable2024} and Jaxley \cite{deistlerJaxleyDifferentiable2025} have incorporated gradient-based learning for certain detailed neuron models, their designs overlook software–hardware co-design and architecture-aware optimization, resulting in inefficient multiscale modeling and limited practicality for deployment on neuromorphic hardware. We demonstrate in Supplementary Fig.~\ref{fig:bench_hh_t} and Fig.~\ref{fig:bench_hh_long} a substantial performance gap ranging between approximately 50 -- 3,000\,$\times$, positively correlated with the sequence lengths.

Furthermore, our architectural philosophy adheres to the principle of "avoid reinventing the wheel". 
Previously, to fully unleash the performance of the rapidly evolving AI ecosystem, applications needed careful, specialized low-level optimization\cite{daoflashattention22023}, which requires specialized skills and experiences in hardware architectures and algorithms. For solutions, the AI community has developed high-level frameworks like Triton\cite{tilletTriton2019} to package fast-iterating hardware. However, many efficiency-oriented neuron modeling frameworks rely on domain-specific low-level implementations\cite{gewaltigNESTNeuralSimulation2007, hinesNEURONSimulation1997, stimbergBrain2IntuitiveEfficient2019}. The negative effect of this approach is the difficulties in development, maintenance, iteration, and migration. In comparison, by grounding on Triton, our platform isolates the engineering concerns about specific hardware architectures, drivers, and middleware from the algorithm implementation. This technical routine can significantly reduce the cost of maintenance and iteration, lowering the engineering barrier for research. 

By synthesizing these capabilities, BrainFuse is expected to initiate a new era of bio-inspired intelligence that exceeds the restrictions of computing tools and further explores the dynamics and morphology. Studies have shown that dendrite structures and integration dynamics are crucial for functioning biological brains\cite{fischerDendriticMechanismsVivo2022, gencDiffusionMarkersDendritic2018, poiraziIlluminatingDendriticFunction2020}. 
Using BrainFuse, neuroscientists can accelerate large-scale network simulation dozens of times and leverage various optimization methods for data-driven discovery. By combining with neuromorphic deployment, they can also evaluate the behavioral functionality of specific neuronal structures and topologies through real-world interactions of an agent. For AI studies, researchers can employ biological dynamics in rather complex and practical learning tasks to explore and verify more applicational advantages\cite{heNetworkModelInternal2024}.
With the support of BrainFuse, we have already discovered multiple interesting properties of complex neuron dynamics. HH-based models are applicable in many gradient-based sequential learning tasks, showing stronger intrinsic robustness to input perturbation than LIF. Joining the efforts of multiple related fields, it is highly possible to uncover the common essence of biological and artificial intelligence. 

To accelerate this interdisciplinary trajectory, we have meticulously engineered the architecture and interface of BrainFuse to be both compact and flexible, suitable for secondary development and interacting with other frameworks. It will be fully released as an open-source platform to foster broad collaboration and invite expert insights from multiple research communities. By maintaining a low barrier to entry, BrainFuse encourages rigorous peer validations, user feedback, and diverse community contributions.
With this open foundation, BrainFuse is expected to fuse the tooling ecosystems of computational neuroscience and AI. Our future roadmap focuses on bridging this divide by directly integrating established simulation protocols (e.g., from NEURON and NEST) with popular AI development frameworks. By consolidating these efforts into an integrated, cohesive workflow, BrainFuse aims to uncover the common essence of biological and artificial intelligence, empowering the next generation of bio-inspired systems.

\section*{Methods}

\subsection*{Biological neural dynamics}

\paragraph{Membrane potential and ion-channel definitions.}
The dynamics of a spatially distributed biological neuron can be described by partial differential equations (PDEs) in Eq.\ref{eq:hh_eq_full}, 
\begin{equation}
\label{eq:hh_eq_full}
    \left\{
    \begin{array}{c@{\hspace{2pt}}c@{\hspace{2pt}}l}
    I &=& C_m \frac{\partial V}{\partial t} + 
        \frac{1}{r_l}\frac{\partial ^ 2V}{\partial x^2} + 
        \sum_X \bar g_X \eta_X(\vec p)(V-E_X) \\
    \frac{\mathrm{d} p}{\mathrm{d} t} &=& \alpha_p(V)(1-p) - \beta_p(V)p
    \end{array}
    \right.
\end{equation}

where $I$ is external stimulus current injection into the neuron, $V$ is membrane potential, $C_m$ is the membrane capacity, $x$ is the position along the fiber, $r_l$ is the per unit length resistance of the neuroplasm, $\bar{g}_X$ is the maximum conductance of ion X, $E_X$ is the equilibrium potential of ion $X$, $p$ is the gating variable, the probabilities of activation of the sub-units in the corresponding ion-channel protein, $\alpha_p$ and $\beta_p$ are the $V$-dependent state transition rates of sub-unit $p$, $\eta_X(\vec p)$ describes how the subunit(s) denoted with $\vec p = [p_1, [p_2, \dots]]$, the activation probability of each subunit, influence the conductance of the channel protein of ion $X$. The specific form of $\eta_X(p)$, $\alpha_p(V)$ and $\beta_p(V)$ depends on the corresponding ion $X$ and sub-unit $p$. Details of model parameters are provided in Supplementary Section \ref{ssec:neuron_definition}.

\paragraph{Analysis of $C_m$.}

In HH neurons, the sensitivity to $I$ is adjustable by tuning $C_m$. For the simplicity of analysis, single point explicit Euler difference is adopted here. The update of $V$ is

\begin{equation}
\Delta V = \frac{\mathrm{d} V}{\mathrm{d}t} \Delta t
         = \frac{\Delta t}{C_m}\left[I+\sum_x g_x(V,t)(V-E_x)\right]
         = \frac{\Delta t}{C_m}\left[I+f(V,t)\right].
\label{eq:hh_dyn_simplified}
\end{equation}

In learnable networks, $I$ is generated by the parameters of the former layers, making $I$ adaptive to the value of $C_m$ guided by the loss function. The second term $f(V,t)$ is determined by the intrinsic dynamics of the equation. As a result, tuning $C_m$ will change the weight ratio between $I$ and $f(V,t)$. Larger $C_m$ suppresses the effect of neuron dynamics and increases the influence of external input. In visualizing the neuronal behavior with or without input noise, we recorded the neurons' membrane potential in a trained MLP network with SHD test set data as clean input.

\subsection*{Implementation and acceleration}

\paragraph{Simplified discretization.}

We took the HH point neuron model\cite{hodgkinQuantitativeDescriptionMembrane1952} as an example of the general form in Eq.\ref{eq:hh_eq_full}. We adopted the exponential Euler method to discretize the temporal ion-channel terms in the HH equations. For the spatial term, we applied the second-order Euler explicit difference. This design can significantly lower the computational complexity while preserving most of the neuronal dynamics, although it would be theoretically less accurate than more precise implicit methods with higher orders, such as the Runge-Kutta method. Nonetheless, higher arithmetic complexity increases the numerical precision loss, especially for single-precision computation commonly used in GPUs. This would be more critical in gradient training scenarios due to the doubled computation path in back propagation through time (BPTT)\cite{werbosBPTT1990}. The detailed discretization scheme is described in Supplementary Section \ref{ssec:discretization}.

\paragraph{Gradient learning support.}

After discretization, the HH functions can be numerically solved with numerical integration or discretized state iteration. We first implemented the state update function in the PyTorch framework. With the auto-gradient feature of PyTorch, arithmetic operations are automatically recorded to construct a gradient computation graph, which intuitively supports back propagation (BP) in deep learning. However, the native implementation with PyTorch is significantly slower than LIF neurons presented in SpikingJelly\cite{fangSpikingJellyOpensourceMachine2023}. We addressed this issue with comprehensive and carefully designed optimizations. The derivation of gradient computation is illustrated in Supplementary Fig.~\ref{fig:hh_comp_graph}.

\paragraph{Engineering optimization.}

GPUs are well-suited for scaled computing tasks. To better fit our platform into modern AI computing infrastructure, we conducted targeted optimization for the characteristics of GPUs. Memory access and kernel launching are overhead-costly procedures. We applied operator fusion and a re-computation strategy to eliminate avoidable memory access. Operator fusion also reduces the amount of kernel launches in the same computation procedure (Supplementary Fig.~\ref{fig:op_fusion}). A comprehensive treatment of the re-computation strategies and memory–compute trade-offs underlying our optimized HH operator can be found in Supplementary Section \ref{ssec:compute_strategy}. We visualized the memory usage reduction in Supplementary Fig.~\ref{fig:mem_comp}. The optimized HH operators are implemented using Triton\cite{tilletTriton2019}, including the state update function and the manually derived corresponding gradient functions. 

\paragraph{Flexibility and extensibility.}

In-depth engineering optimization usually increases the complexity of the platform, thus impeding flexible customizations. We avoided this dilemma by leveraging the AI ecosystems. The complicating part of the HH computation is mainly parallel processing, gradient graph composition, and low-level interface interactions, while the definition of the dynamics equations are straightforward. Our platform packages the cumbersome part and exposes the definition of ion-channel dynamics with Triton syntax. Following this design, users can easily define new dynamics in an intuitive form while enjoying consistent efficiency.

\paragraph{Neuromorphic deployment.}

We adopted a neuromorphic chip with a heterogeneous architecture that merges bio-inspired computing with general-purpose computing. 
To deploy the cortical network simulation algorithm written in Python, the code needs to be converted to equivalent C/C++ code to adapt to the architecture of the neuromorphic chip. Next, chip-specific adaptations are applied to fully leverage the chip's features for higher performance and lower power consumption. The adaptation includes replacing the standard C library's memory management interface with the chip's dedicated version and optimizing spike delivery mechanisms. Finally, the modified code is compiled with a customized GCC compiler designed for the neuromorphic chip. 
We evaluated the speed and power consumption with the customized GEM5 simulation.

\subsection*{Gradient-based learning tasks}

\paragraph{Training and preprocessing.}

In most gradient-based learning tasks, we adopted the same network structure and hyperparameters as the compared works, Adam or AdamW optimizer, CosineAnnealing or chained CosineAnnealingLR, and LinearLR scheduler for training. Sequential image data is generated by flattening the 2D image into a 1D pixel sequence. For colored images, the channel dimension is preserved. See Supplementary Section \ref{ssec:ML_config} for detailed configurations.

\paragraph{Robustness evaluation.}

Input noise robustness is evaluated by training the model on the original clean training dataset and then testing on the noisy testing set. In Fig.~\ref{fig:robust}a, we applied different kinds of noises according to the properties of the data modality. Specifically, we adopted the CIFAR10-C dataset, a specialized 2D image dataset for perturbation robustness. The 2D images are flattened as other image data. Detailed noise configurations and additional results are shown in Supplementary Table \ref{table:noise1}, \ref{table:noise2}, and Supplementary Fig.~\ref{fig:robust_curve}.

\subsection*{Biological modeling}

\paragraph{Symmetric mean average percentage error (sMAPE).}

sMAPE is used to measure the relative goodness of fitting a curve with a trainable neural network. Mean Average Percentage Error (MAPE) is defined as:
\begin{equation}
    \operatorname{MAPE}(\vec y,\hat {\vec y}) = 100\% * \frac1n \sum_{i=0}^{n} \left| \frac{\hat y_i - y_i}{y_i} \right|,
\end{equation}

where $\hat {\vec y}$ is the ground truth value vector, $\vec y$ is the predicted value vector, the subscript $i$ represents the scalar value of the i-th element in the vector, and $n$ is the total count of the tested samples. However, MAPE has an undefined value issue at $y_i = 0$ when predicted $y_i$ is 0. sMAPE is a modified version of MAPE to resolve this issue:
\begin{equation}
    \operatorname{sMAPE}(\vec y,\hat {\vec y}) = 100\% * \frac1n \sum_{i=0}^{n} 
    \frac{\left|\hat y_i - y_i\right|}{|\hat y_i| + |y_i|}.
\end{equation}
sMAPE ranges within $[0, 1]$ and smaller sMAPE indicates better fitting. We adopted the Mann-Whitney test to justify the significance of the comparisons. 

\paragraph{L5PC fitting.}

The L5PC dataset is generated using NEURON numerical simulation of a finely modeled biologically realistic neuron with multiple ion channel dynamics and morphology. The model contains 639 input synapses, one excitatory and one inhibitory channel for each synapse, and a soma. The recorded data contains input spikes at all synapse channels, soma membrane potential, and spikes. The test split of the L5PC dataset is divided into different simulation setup groups: standard, sub-threshold, out-of-distribution, and Ergodic. The train split is all generated by the ergodic setup. 

We applied an end-to-end fitting with a single HH1952 neuron. The dendrites are modeled by a single linear layer. We added a 1D temporal convolution layer before the dendrites using a manually designed kernel to simulate the E/IPSP filters.  For optimization, we use MSE loss for membrane prediction and CrossEntropy loss for spike prediction. The learning rate is $5\times10^{-4}$. A CosineAnnealing scheduler is applied to tune the learning rate.

\paragraph{C. elegans sensing-motion circuit.}

The C. elegans dataset summarized various C. elegans neural activity recording data into a comprehensive dataset. The recordings are Ca imaging data, containing temporally aligned cell activity traces of neurons inside C. elegans. However, due to limitations in experimental procedures, different labs and trials have different tracked neurons. For consistency, we chose 18 from the 203 neurons (L/R for each): ASI, ASE, AFD, AWA, AWB, AWC as sensory input and AVA, AVB, AVE for motion command output\cite{tsalikFunctionalMappingNeurons2003}.

The recordings are cut into segments of the same length (10ms) for training. As the neural circuit is stateful, we preserved a 20ms segment without supervision at the beginning of each sample. In order to yield the most sample volume as possible, we overlapped the unsupervised segments in segmenting input signals, as shown in Fig.~\ref{fig:c_elegans} (\textbf{c}). The segmented signals are composed of 12ms input signal and 10ms output signal pairs. Following this process, 4,694 samples were generated and randomly partitioned into training and testing sets using a 3:1 ratio.

\paragraph{Cerebral cortex network.}

Tobias C. Potjans and Markus Diesmann proposed a cerebral cortex connectivity model based on anatomical and physiological measurement data. We constructed a sparsely connected reservoir with BC2008 dynamics\cite{pospischilMinimalHodgkinHuxley2008} spiking neurons following the reported connection model. Due to different dynamics, the parameter tuning rule reported in \cite{potjansCell-type2014} is not applicable. Thus, the connection strength average and standard deviation are empirically configured. For computational optimization, we deduced the form of distribution of the background input after integration, which is $\sum_i \text{Poisson}(\lambda)_i \cdot \text{Gaussian}(\mu, \sigma)_i $ (see details in Supplimentary Section \ref{ssec:cortex_input_simp}), and directly generate samples under this distribution without computing the component values. The other configurations, including E/IPSP filters and synaptic delays, are kept identical as reported (see Supplementary Section \ref{ssec:cortex_engineering}). The simulation step is 0.1ms. 

For rest-state simulation, we adopted mean=$0.26~\mathrm{\mu A}$, std=$0.026~\mathrm{\mu A}$ for recurrent connection strength, mean=$0.17~\mathrm{\mu A}$, std=$0.017~\mathrm{\mu A}$ for background connection strength, and background poisson rate=$8$Hz. 

For thalamo-cortical transient stimulation, we adopted mean=$0.26~\mathrm{\mu A}$, std=$0.026~\mathrm{\mu A}$ for recurrent connection strength, mean=$0.22~\mathrm{\mu A}$, std=$0.022~\mathrm{\mu A}$ for background connection strength, and background poisson rate=$4$Hz.

\subsection*{HH encryption}

We used Look-Up Table (LUT) mapping to perform encoding and Cipher Block Chaining (CBC) to improve security. The complete procedure of the HH encryption-decryption algorithm is as follows.

A complete encryption-decryption algorithm contains 2 mappings, implemented with a mapping LUT A and an inverse mapping LUT B, following the constraint $\forall i \in[0, n) \cap \mathbb{N}, B[A[i]]= i$, where $[\cdot]$ represents array subscription. During encryption, the input bytes are chunked into a sequence of segments that can be treated as $n$-bit unsigned integers. Use the integer as the index to fetch from LUT A will get the corresponding ciphertext. Conversely, during decryption, the inverse map LUT B is applied.

The required invert map B is implicitly represented by the order of $(\mathrm{f}, \mathrm{A})$ pairs (the key), i.e., aligning the domain and range of LUT B with the key index and the $n$-bit response combination of the HH filter bank:
$[\mathrm{Downsample}\circ \mathrm{HH}|_{\tau, \cdots}](key[i]) = \operatorname{dsHH}(key[i])$
with invert LUT B, or 
$\forall k \in[0, n) \cap \mathbb{N}, \operatorname{dsHH}(\operatorname{key}[k])=B[k]$
, or 
$\operatorname{dsHH}(key[A[k]])=k$.

The response of the HH filter bank is jointly determined by signal parameters (f, A) pairs and HH parameters, including $\tau$, ion channel equations, etc. These configurations are all necessary for correct decryption and can be used either respectively or jointly as the secret key. The (f, A) pair sequence and $\tau$ combination of the HH filter bank are treated as private and public keys in the demo experiment. The decryption result degenerated under tiny perturbation following normal distribution $N(0,10^{-12})$. A detail analysis of HH's frequency selectivity and description of the algorithm are provided in Supplementary Section \ref{ssec:hh_encryption}.

\section*{Data availability}
Datasets for gradient-based learning are available in referenced works. Codes and experiment results are available upon request. The code of our framework will be released upon acceptance.

\section*{Acknowledgments}
We thank Yuanhong Tang, Qingyan Meng, and Liutao Yu for the patient discussion on the setup details of cortical network simulation; we thank Yuhong Chou for the intuitive technical discussion on triton optimization; we thank Yunhui Xu and Linxuan He for the assistance in the early stage exploration of HH dynamics modeling; we thank Xia Liu, Jin Bao, and Botong Liu for the careful and professional revision of our manuscript; we thank Boyang Ma, Yunhao Ma, Chenlin Zhou, and Sihang Guo for the thoughtful discussion on paper writing and visualizations.

\section*{Author contributions}
B.C. developed the BrainFuse platform, performed gradient-based learning, biological modeling experiments, and data analysis, wrote the paper; Y.W. and B.X. participated in paper revision; S.X. participated in data analysis, and paper revision; P.Q., D.W., X.C. and H.B. deployed the cortical network on the neuromorphic chip; S.Z. participated in background investigation and paper revision; G.L., Z.M., Y.W., and Y.Z. supervised the work. All authors discussed the results and reviewed the manuscript.

\section*{Competing interests}
The authors declare no competing interests.

\lstset{
    language=Python,
    basicstyle=\footnotesize\bfseries\ttfamily,
    commentstyle=\color[HTML]{228B22},
    emphstyle=\color[HTML]{000088},
    stringstyle=\color[HTML]{228B22},
    numberstyle=\color[HTML]{006666},
    tabsize=2,
    keywordstyle=\ttfamily,  
    xleftmargin=0.5cm, 
    emph={Sequential,Linear,ReLU,HHNode,jit,tensor,exp},
    showspaces=false,      
}

\newcommand{\Plus}{\makebox[0pt][l]{\color[HTML]{d7e7b3}\rule[-4pt]{0.99\linewidth}{14pt}}}
\newcommand{\Mod}{\makebox[0pt][l]{\color[HTML]{ebf1dd}\rule[-4pt]{0.99\linewidth}{14pt}}}
\newcommand{\Minus}{\makebox[0pt][l]{\color[HTML]{ffcccc}\rule[-4pt]{0.99\linewidth}{14pt}}}




\newpage

\bibliography{ref_list}

\renewcommand{\thesection}{S\arabic{section}}  
\renewcommand{\thetable}{S\arabic{table}}  
\renewcommand{\thefigure}{S\arabic{figure}}
\renewcommand{\figurename}{Figure}
\setcounter{figure}{0} 
\setcounter{table}{0} 

\addtocontents{toc}{\protect\setcounter{tocdepth}{-1}}
  
\end{document}